\documentclass{article}

\usepackage{arxiv}
\usepackage[utf8]{inputenc} 
\usepackage[T1]{fontenc}    
\usepackage[colorlinks=true,linkcolor=blue,citecolor=blue,urlcolor=blue]{hyperref} 
\usepackage{url}            
\usepackage{booktabs}       
\usepackage{amsfonts}       
\usepackage{nicefrac}       
\usepackage{microtype}      
\usepackage{lipsum}         
\usepackage{graphicx}
\usepackage{natbib}
\usepackage{doi}
\usepackage{cite}
\usepackage{multirow}
\usepackage{tabularx}
\usepackage{amsmath,amssymb}
\usepackage{algorithmic}
\usepackage{textcomp}
\usepackage[table,xcdraw]{xcolor} 
\usepackage{colortbl} 

\definecolor{lightblue}{rgb}{0.8,0.85,1}
\definecolor{lightyellow}{rgb}{1,1,0.85}
\definecolor{lightblue1}{rgb}{0.8,0.85,1}
\definecolor{lightyellow1}{rgb}{1,1,0.85}

\title{Machine Learning-based Android Intrusion Detection System}

\author{
Madiha Tahreem \\
Z.H. College of Engineering and Technology \\
Aligarh Muslim University \\
Aligarh, India \\
\texttt{madihatahreem487@gmail.com} \\
\And
Ifrah Andleeb \\
Z.H. College of Engineering and Technology \\
Aligarh Muslim University \\
Aligarh, India \\
\texttt{ifrahzhcet@gmail.com} \\
\AND
\href{https://orcid.org/0009-0008-9399-9103}{\includegraphics[scale=0.06]{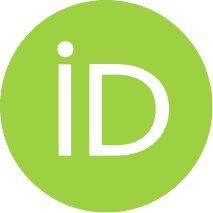}\hspace{1mm}Bilal Zahid Hussain} \\
Z.H. College of Engineering and Technology \\
Aligarh Muslim University \\
Aligarh, India \\
\texttt{zahidhussain909@gmail.com} \\
\And
Arsalan Hameed \\
Z.H. College of Engineering and Technology \\
Aligarh Muslim University \\
Aligarh, India \\
\texttt{arsalanhameed@zhcet.ac.in} \\
}

\renewcommand{\shorttitle}{Intrusion Detection System}

\hypersetup{
pdftitle={A template for the arxiv style},
pdfsubject={q-bio.NC, q-bio.QM},
pdfauthor={David S.~Hippocampus, Elias D.~Striatum},
pdfkeywords={First keyword, Second keyword, More},
breaklinks=true
}

\begin{document}
\maketitle

\begin{abstract}
The android operating system is being installed in most of the smart devices. The introduction of intrusions in such operating systems is rising at a tremendous rate. With the introduction of such malicious data streams, the smart devices are being subjected to various attacks like Phishing, Spyware, SMS Fraud, Bots and Banking-Trojans and many such. The application of machine learning classification algorithms for the security of android APK files is used in this paper. Each apk data stream was marked to be either malicious or non malicious on the basis of  different parameters. The machine learning classification techniques  are then used to classify whether the newly installed applications’ signature falls within the malicious or non-malicious domain. If it falls within the malicious category, appropriate action can be taken, and the Android operating system can be shielded against illegal activities.
\end{abstract}


\section{Introduction}

The Android smartphone operating systems currently at 85\% occupy one of the largest market spaces, including mobile phones, tablets, and even smartwatches. Given their presence in the market, one would believe they are relatively immune to malware and malicious activities but that does not seem to be the case. Unlike Apple’s iOS, Android Operating Systems are open source which gives ample opportunity to hackers to use and create their own malware and get hold of the user’s sensitive information. Java is used to build Android applications which are then fabricated by data and resource files into an archive known as Android Package Kit(APK). APKs are responsible for the source code, assets and resources integral for running an application. There are currently more than 970 million pieces of malware circulating the internet right now(AV-test institute). In a recent report, Cisco Umbrella received a record-high 96.39 percent threat detection rate. The sheer number of samples has exhibited the manual analysis, and classification as majorly impractical, and this, in turn, has made cybersecurity\citep{sabir2021} researchers, mobile app store holders, and antivirus companies automate the threat recognition process. On top of that, it is estimated that about a whopping 47\% of mobile anti-malware apps fail to detect critical threats. Every threat in an android smartphone is related to a particular dimension such as privacy, data and communication. Some of these security challenges include Phishing, Spyware, SMS Fraud, Bots and Banking-Trojans. It has been claimed that Download-Trojan apps download their harmful code after installation, making it difficult for Google technology to readily identify them. Malware attacks are done through the permission mechanism and untrusted APKs\citep{maligu} and include even simple web browsing which ends up in installing a malicious code on the device in backend; Privacy attacks including leakage of a user’s confidential data, financial details and social media platform’s credential. These are usually done through means of suspicious SMS and emails in which the user may end up losing sensitive information. These results in cyberattacks making money illegally. Physical attacks consist of recognising the password pattern of a device through means of touch screen. 
The existing antivirus tools can only help against known malware\citep{Mira2021ASL}, implying that unrecognized or newly developed malware can go unnoticed for months. To tackle this, machine learning\citep{9673587} algorithms in android malware detection systems have proven to be immensely useful in identifying malware through behavior-based and anomaly based software analysis. 

\textit{Contribution}:  The main contributions of this paper can be summarized as follows: 

(i) Utilize the various machine learning classifiers to detect behavior using features from an Android device, majorly the permissions required by an APK file to detect malicious activity, mainly monitoring Binder API features, install-time permissions and runtime permissions. 

(ii) Discuss the best classier methodology based on performance metrics; Random Forest Classifier[9] and discuss the number of trees in the forest and the amount of random features chosen for each tree node are varied for Android-based features, assess the accuracy of the Random Forest classifier. 

The remainder of the paper is structured as follows. Related works are described in Section II. Section III gives an overview about proposed methodology which includes a brief about the dataset and a detailed view on the methodology approach.

\section{RELATED WORK}
\label{sec:rel}
Machine Learning is extensively used to perform various tasks such as object and text identification and recognition, feature extraction. Moreover, it is also being used in detection of malicious Android applications. 

\citet{10.1155/2021/9964224} in their research work proposed a deep network fusion model-based strategy for detecting malicious Android applications. The technique extracts the static core code features by decompiling APK files. It then vectorizes the code and applies a deep learning network for classification and descrimination. 

\citet{10.1145/3293475.3293489} in their paper validated two Machine Learning algorithms, K-Nearest Neighbors (KNN) and Support Vector Machine (SVM) to conduct supervised learning based classification of the feature set into either hazardous or benign applications. The experimental results show average accuracy rates for a dataset of realistic malware and benign applications are 80.50\% and 79.08\% respectively.

\citet{10.1007/978-3-319-50127-7_11} also deployed Convolutional Networks, long-term and short-term memory networks to evaluate the calling sequence in their research work. Although, the feature information is insufficient, and they merely model and classify harmful call sequences

\section{METHOD}
\label{sec:met}
To stop attacks that evolve daily, intrusion detection systems must be regularly updated. Anomaly-based network intrusion detection systems are now a commonly used technology in the attack detection stage in this context. We chose a variety of machine learning techniques for this use-case since machine learning may be used to successfully achieve outstanding accuracy and precision. The dataset used, the suggested operational framework, the proposed network's design, and the evaluation metrics are all covered in the parts that follow.

\subsection{Dataset}
\begin{figure}[ht]
\begin{center}
\includegraphics[width=0.7\textwidth]{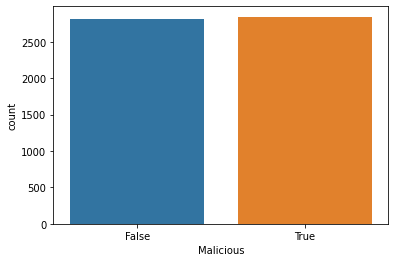}
\caption{Breakdown of the MRI dataset for different classes across the training and testing subsets}
\label{fig:mri_data}
\end{center}
\end{figure} 
The proposed machine learning models are trained and evaluated using an open-access benchmark dataset called “Deep Learning for Cyber Security\footnote{https://www.kaggle.com/competitions/deep-learning-for-cyber-security/overview}”, which contains 8078 samples. The dataset shows a collection of APK files together with their permissions. Fig. 1 depicts the distribution of malicious and non-malicious samples. We divided the dataset 70:30 for testing and training our suggested model. The dataset used to train the suggested machine learning models comprises 4039 malicious and 4039 non-malicious samples.

\subsection{Evaluation Metrics}

\paragraph{Accuracy} One way to gauge how frequently a machine learning classification algorithm classifies a data point correctly is to look at the algorithm's accuracy. The proportion of accurately predicted data points among all the data points is known as accuracy. The number of true positives and true negatives divided by the total number of true positives, true negatives, false positives, and false negatives is how it is more precisely described. A data point that the algorithm correctly identified as true or untrue is referred to as a true positive or true negative. On the other hand, a false positive or false negative is a data item that the algorithm misclassified.

\paragraph{Precision} Precision, or the caliber of a successful prediction made by the model, is one measure of the model's performance. Precision is calculated by dividing the total number of positive predictions by the proportion of true positives (i.e., the number of true positives plus the number of false positives). For instance, in a customer attrition model, precision is the ratio of the total number of consumers the model properly anticipated would unsubscribe to the number of customers who actually did so.

\paragraph{Recall} The recall is determined as the proportion of Positive samples that were correctly identified as Positive to all Positive samples. The recall gauges how well the model can identify Positive samples. The more positive samples that are identified, the larger the recall. Only the classification of the positive samples is important to the recall. This is unrelated to the manner in which the negative samples are categorized, such as for precision. Even if all of the negative samples were mistakenly classified as Positive, the recall will be 100\% when the model labels all of the positive samples as Positive.

\subsection{Methodology and approach}
In the study, Accuracy, Precision and Recall parameters were calculated by running iterations of SVM, Random Forest, Linear Discriminant Analysis and Light GBM classifiers. 

\paragraph{Support Vector Machines(SVM) \citep{Boser1992ATA}} The interval maximization in the eigenspace is defined by a linear classifier as the basic model of the two-class SVM. The learning strategy of SVM is to maximize the interval, which can be formalized as a problem of solving convex quadratic programming. This technique is also referred to as the maximum edge algorithm, and it has the advantage of having a strong generalization ability that allows it to address problems such as nonlinearity, small sample size, high dimension, etc. Using the linear separable SVM as an example, the principle of SVM is to find a separable hyperplane in a particular eigenspace and then split the sample space into two categories: a positive class and a negative class, which correspond to two different types of samples.

\paragraph{Random Forest \citep{random}} Random Forest is an ensemble learning algorithm. An approach known as an ensemble learner produces numerous individual learners and aggregates the outcomes. An improvement on the bagging method is used by random forest. In a typical decision tree, each feature characteristic is taken into consideration while making a decision at a node split. However, in Random forest, a random selection of features is used to choose the optimal parameter at each node of a decision tree. In addition to helping Random Forest scale effectively when there are numerous features in each feature vector, this random selection of features also makes it less susceptible to data noise and reduces the interdependence (correlation) between the feature qualities.

\paragraph{Linear Discriminant Analysis \citep{tharwat}} It is a fairly straightforward probabilistic classification model that generates reliable predictions even when its strong assumptions about the distribution of each input variable are broken. A probabilistic model for each class is created using the technique, based on the unique distribution of observations for each input variable. Consequently, in classifying a new example is to determine the conditional likelihood that it belongs to each class before choosing the one with the highest probability.

\paragraph{Light GBM \citep{Ke2017LightGBMAH}} A gradient boosting framework that uses tree based learning algorithms. While other algorithms grow trees horizontally, Light GBM grows vertically, which translates to Light GBM growing trees leaf-wise while other algorithms grow level-wise. The leaf with the greatest delta loss will be chosen to grow. It can reduce loss more than a level-wise approach when expanding the same leaf. Light GBM is preferred since it uses less memory,  can handle even large amounts of data and its emphasis on precise outcomes.

\section{Results \& Discussion}
\label{sec:res}
Table 1 compares various approaches such as SVM, Random Forest, LDA and Light GBM for Android malware detection. It is clearly evident from Table 1 that results from all other methods show lower accuracy than Random Forest. All the algorithms are trained for the same “Deep Learning for Cyber Security” Dataset for uniform comparisons.

The confusion matrix based heatmaps of various algorithms are given in Fig 2, Fig 3, Fig 4 and Fig 5.
\begin{table}[ht]
\centering
\label{table:acc_table}
\caption{Comparison table of different Machine Learning Algorithms}
\begin{tabular}{|c|c|c|c|}
\hline
\textbf{Method}        & \textbf{Accuracy(\%)} & \textbf{Recall(\%)} & \textbf{Precision(\%)} \\ \hline
\textbf{SVM}           & 93.81                 & 93.88               & 93.89                  \\ \hline
\textbf{Random Forest} & 99.11                 & 99.53               & 99.88                  \\ \hline
\textbf{LDA}           & 95.05                 & 94.37               & 95.71                  \\ \hline
\textbf{Light GBM}     & 97.76                 & 98.00               & 98.00                  \\ \hline
\end{tabular}
\end{table}
\begin{figure}[ht]
\begin{center}
\includegraphics[width=0.5\textwidth]{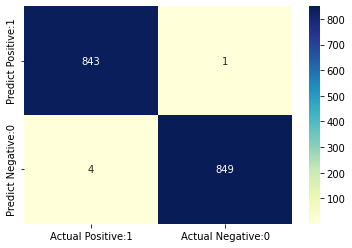}
\label{fig:rf}
\caption{Visualization of confusion matrix with heatmap of Random Forest}
\label{fig:class_arch6}
\end{center}
\end{figure} 
Random Forest provides an exceptionally high accuracy, recall and precision of 99.11\%, 99.53\% and 99.88\% respectively which proves the efficiency of this algorithm. Random Forest is set for 10 folds, 10 N\_splits and 140 N\_estimators. Maximum depth is taken to be 22 for optimal results. All the hyperparameters are tuned by using GridsearchCV. The top 10 focus features for the Random Forest classifier are shown in Fig. 6.

\begin{figure}[ht]
\begin{center}
\includegraphics[width=0.5\textwidth]{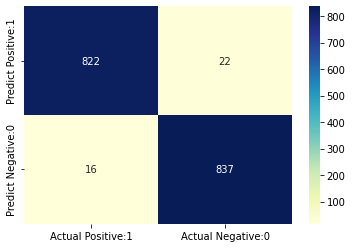}
\label{fig:lgbm}
\caption{Visualization of confusion matrix with heatmap of Light GBM}
\label{fig:class_arch1}
\end{center}
\end{figure} 
Light GBM also shows promising results with an accuracy of 97.76\%, a recall and precision of 98.00\% each for detecting malware in android applications. The True Positives or Predict Positives are found to be 822 and can be visualized from  Figure 3. 

\begin{figure}[ht]
\begin{center}
\includegraphics[width=0.5\textwidth]{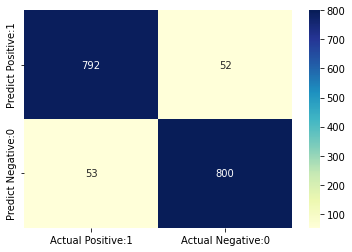}
\label{fig:svm}
\caption{Visualization of confusion matrix with heatmap of SVM}
\label{fig:class_arch2}
\end{center}
\end{figure} 
\begin{figure}[ht]
\begin{center}
\includegraphics[width=0.5\textwidth]{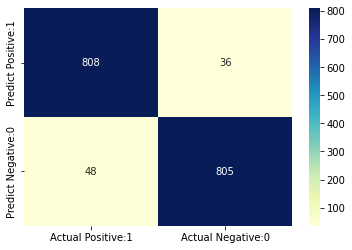}
\label{fig:lda}
\caption{Visualization of confusion matrix with heatmap of Linear Discriminant Analysis}
\label{fig:class_arch3}
\end{center}
\end{figure} 
\begin{figure}[ht]
\begin{center}
\includegraphics[width=0.6\textwidth]{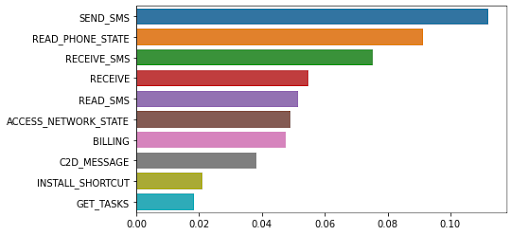}
\caption{Visualization of 10 most important features for Random Forest classifier}
\label{fig:class_arch4}
\end{center}
\end{figure} 
We can analyse the following about the Android features under consideration based on our experimental findings from 10-fold cross validation:
\paragraph{High Accuracy of Random Forest} Random forest has an unusually high accuracy, properly classifying over 99.1\% of the data. The Recall and Precision are 99.53 and 99.88 respectively. With n\_estimators=140, max\_depth=22 and randomly chosen features were the best configuration. 
\paragraph{Misclassified malicious samples} Across all tests, the majority of misclassified cases resulted from malicious class samples being incorrectly categorized as benign class samples. This is based on looking at the confusion matrix that is generated for each experiment condition. Although the present dataset produces some extremely encouraging findings, its current form makes it difficult to use for real-time monitoring of malware infestations on Android devices. The existing detection method would show promise in the case where the Android.apk executable application file is tested in a private setting before being installed on a real device or before being made available to the general public. But this doesn't happen very often. We must comprehend that the features found in the dataset are of a global nature.

\section{Conclusion}
\label{sec:conc}
Numerous classification techniques from machine learning have been used, as detailed in this work, to determine the best method for identifying malware infections on mobile devices. On mobile devices, both signature-based and behavior (Anomaly)-based techniques are being tested, much like when detecting malware infestations on PCs. In this research, we investigate if and to what extent malware infections can be detected using the machine learning classifiers SVM, Random Forest, Linear Discriminant Analysis, and Light GBM. The dataset was automatically created by keeping track of a collection of 70 APK permission attributes. Random Forest provided an extraordinarily high accuracy of over 99.1\% of the samples accurately identified in our trials, which were based on a 10-fold cross validation.

\bibliographystyle{unsrtnat}
\bibliography{references}

\end{document}